\documentclass[10pt,twocolumn,letterpaper]{article}

\usepackage{cvpr}              % To produce the CAMERA-READY version
\usepackage{algorithm}
\usepackage{algorithmic}

\usepackage{mathrsfs}

\usepackage{colortbl}
\usepackage{color}
\usepackage{multirow}
\usepackage{xr}

\usepackage{xcolor}
\usepackage{rotating}

\definecolor{cvprblue}{rgb}{0.21,0.49,0.74}
\usepackage[pagebackref,breaklinks,colorlinks,allcolors=cvprblue]{hyperref}

\def\paperID{7738} % *** Enter the Paper ID here
\def\confName{CVPR}
\def\confYear{2026}

\title{Rethinking Parameter Sharing as Graph Coloring for Structured Compression}

\author{Boyang Zhang$^{1,2,3}$, Daning Cheng$^{1}$\thanks{Corresponding author}, Yunquan Zhang$^{1}$\\
    $^{1}$Institute of Computing Technology, Chinese Academy of Sciences, Beijing, China\\
    $^{2}$University of Chinese Academy of Sciences, Beijing, China \\ $^{3}$Peng Cheng Laboratory, Shenzhen, China}
\begin{document}
\maketitle
\begin{abstract}
Modern deep models have massive parameter sizes, leading to high inference-time memory usage that limits practical deployment. Parameter sharing, a form of structured compression, effectively reduces redundancy, but existing approaches remain heuristic—restricted to adjacent layers and lacking a systematic analysis for cross-layer sharing. However, extending sharing across multiple layers leads to an exponentially expanding configuration space, making exhaustive search computationally infeasible and forming a critical bottleneck for parameter sharing.
We recast parameter sharing from a group-theoretic perspective as introducing structural symmetries in the model’s parameter space. A sharing configuration can be described by a coloring function $\alpha:L\rightarrow C$ (L: layer indices and C: sharing classes), which determines inter-layer sharing groups while preserving structural symmetry.
To determine the coloring function, we propose a second-order geometric criterion based on Taylor expansion and the Hessian spectrum. By projecting perturbations onto the Hessian’s low-curvature eigensubspace, the criterion provides an analytic rule for selecting sharing groups that minimize performance impact, yielding a principled and scalable configuration procedure.
Across diverse architectures and tasks, Geo-Sharing consistently outperforms state-of-the-art heuristic sharing strategies, achieving higher compression ratios with smaller accuracy degradation.
\end{abstract}    
\section{Introduction}
\label{sec:intro}
\begin{figure}[!ht]
\centering
\includegraphics[width=8.3cm,height=2.7cm]{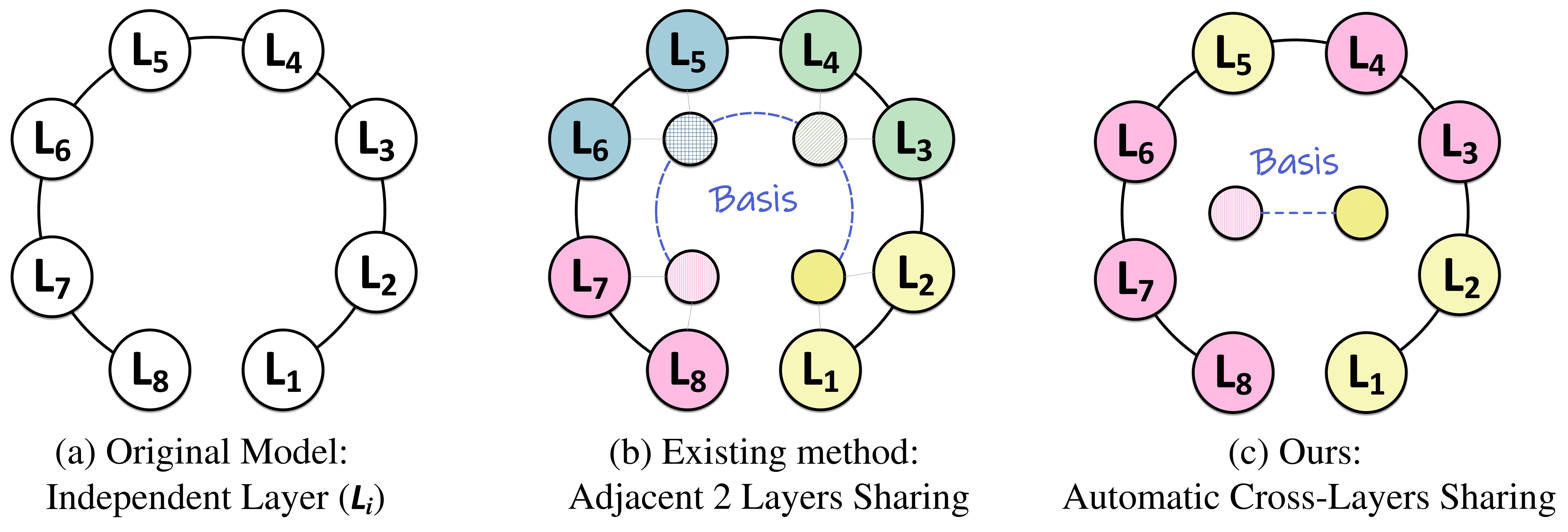}
\caption{Existing methods are heuristic-based and limited to adjacent 2 layers sharing, while our method is theoretically guided to achieve automatic cross-layer sharing and uses fewer basis.}
\label{group}
\end{figure}

As deep models grow increasingly large, the memory and compute overhead at inference time limits their deployment in edge scenarios. Model compression is therefore crucial. A wide range of compression techniques—quantization \cite{zhang2024compression,liu2024spinquant, zhang2025general, zhang2024fp}, decomposition \cite{yuan2025asvdactivationawaresingularvalue, hsu2022languagemodelcompressionweighted, wang2024svd,zhang2024lossless}, and distillation \cite{sun2024logit, moslemi2024survey}—have achieved strong results across settings. However, these methods often require hardware or training pipeline modifications, or incur substantial retraining/fine-tuning costs. By contrast, as a hardware-agnostic and deployment-friendly alternative, weight sharing based on low-rank approximations \cite{lv2023lightformer, 2025learningparametersharingtensor, wang2024basis} is highly attractive: it enables parameter sharing without hardware changes by constructing a low-rank basis for each layer and reusing it across layers.

However, existing sharing strategies largely rely on heuristic rules (e.g., sharing only between adjacent layers \cite{lv2023lightformer, wang2024basis, 2025learningparametersharingtensor} in Figure \ref{group}), which makes it difficult to systematically explore cross-layer similarities. The search space of sharing configurations grows exponentially —for a 32-layer Transformer with 4 shared bases per layer, there are $4^{32}$ possible assignments —rendering exhaustive or heuristic search ineffective, and there is a lack of theoretically grounded, comparable criteria.

\begin{figure*}[!ht]
\centering
\includegraphics[width=16cm,height=6.5cm]{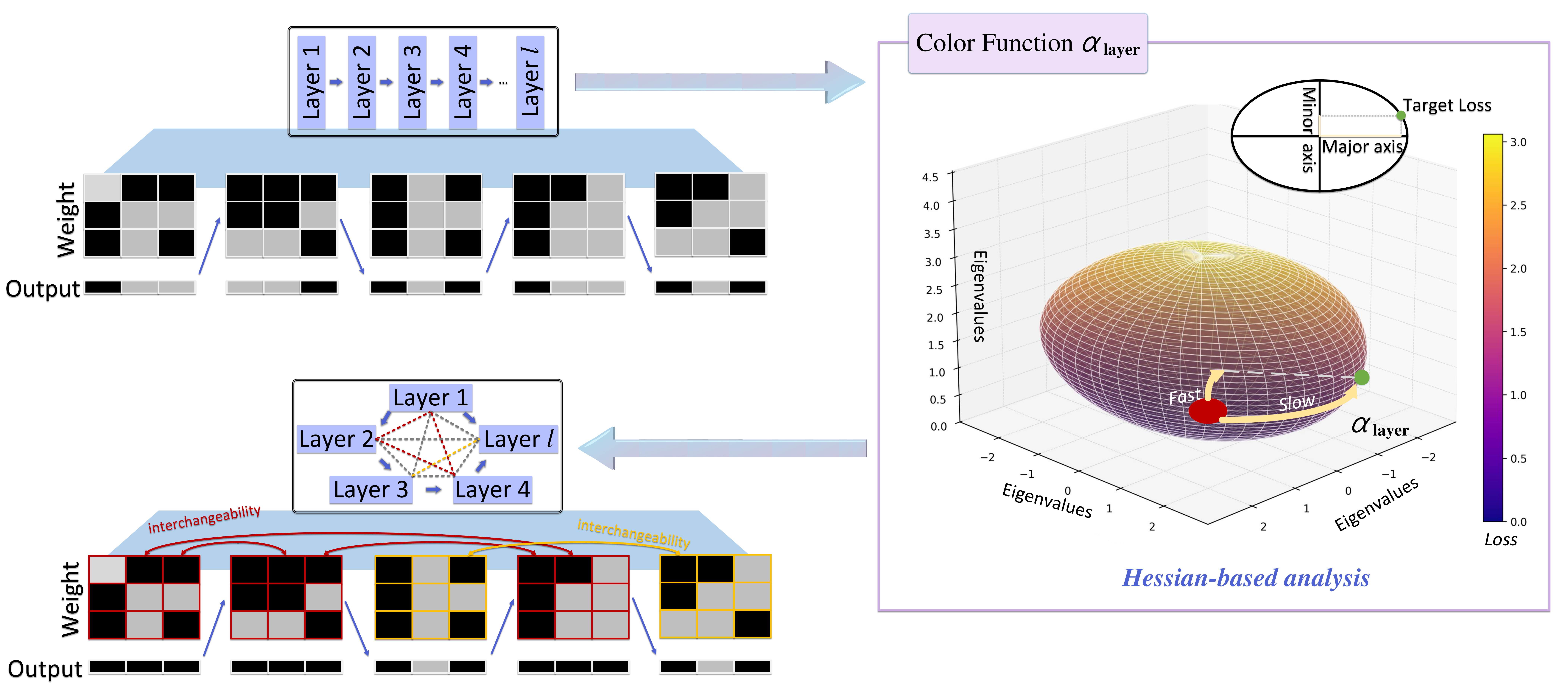}
\caption{Geo-Sharing: The original hierarchical parameter structure is remodeled as a graph, and cross-layer isotropic relationships are achieved through graph coloring. The coloring function $\alpha_{\text{layer}}$ is based on second-order geometric derivation, minimizing the loss growth on the shared error principal axis in the low curvature direction of the Hessian. The right-hand figure shows the alignment effect between the target loss terrain and the coloring rules in the Hessian analysis.}
% \vspace{-0.3cm}
\label{pipeline}
\end{figure*}

Crucially, such heuristic sharing strategies reveal a deeper structural property: when multiple layers can reasonably share parameters, they should be functionally interchangeable—i.e., permuting their order does not fundamentally alter the model’s expressive power. This interchangeability can be rigorously characterized by symmetry in a mathematical sense.
Building on this observation, as shown in Figure \ref{pipeline}, we introduce Geo-Sharing, a theoretically grounded framework for configuring inter-layer sharing. Geo-Sharing aims to address two central questions: what constitutes good sharing (interchangeability) and how to identify it (performance robustness).

For the first question, we formalize inter-layer parameter sharing as introducing structural symmetry in parameter space. Concretely, we use a “coloring function” $\alpha: L\rightarrow C$ to group layers with similar shapes and functions into the same sharing group (same color). Under sharing induced by $\alpha$, layers of the same color use identical parameters and thus remain equivalent under permutations. This sharing-induced interchangeability turns structured compression into the imposition of symmetry constraints. A “good sharing” is a grouping $\alpha$ that satisfies such interchangeability while minimally harming performance.

Symmetry provides the necessary structural condition for sharing, whereas assessing performance robustness requires additional theoretical tools.
To this end, for the second question, we propose a second-order geometric criterion that converts the theoretical notion of “good sharing” into a computable selection rule. Specifically, we view the second-order approximation of the loss as an ellipsoid in parameter space, and align the sharing error subspace induced by $\alpha$ with the low-curvature eigensubspace of the Hessian to minimize the error introduced by symmetry constraints. This alignment admits a closed-form solution via an orthogonal decomposition, turning the selection of a good $\alpha$ into an executable computation.
Empirically, Geo-Sharing preserves accuracy better at high compression ratios on both vision and language models, and outperforms strong SVD-based baselines on generative and downstream inference tasks without any fine-tuning.
Our key contributions are summarized as follows:
\begin{itemize}
    \item We recast inter-layer parameter sharing from a group-theoretic perspective, defining it as the introduction of structural symmetries into the model's parameter space. This formulation transforms sharing from an empirical heuristic into a mathematically principled property of model architecture.
    \item We develop a geometric criterion that determines sharing configurations by aligning the Hessian’s low-curvature subspace with the subspace induced by the coloring function $\alpha$. This criterion transforms the exponentially large configuration search into a closed-form curvature-aligned optimization, providing both analytical interpretability and computational efficiency.
    \item The proposed Geo-Sharing framework enables training-free compression guided by geometric symmetry, achieving superior compression–accuracy trade-offs across vision and language models.
\end{itemize}

\section{Related Work}
\label{sec:formatting}
Since Geo-sharing employs SVD-based weight decomposition to form shared bases, we review the relevant works on SVD-based weight decomposition and parameter sharing.

%-------------------------------------------------------------------------
\subsection{SVD-based Weight Decomposition}

Weight compression via singular value decomposition (SVD) or low-rank approximation is a widely studied technique in neural network model compression. Early approaches \cite{golub1987generalization} proposed decomposing a weight matrix with SVD and retaining only the principal singular values to reduce parameters and optimize storage. In architectures like Transformers \cite{vaswani2017attention}, however, outliers in weights and activations can introduce significant errors during compression \cite{lv2023lightformer, wu2023singularformer}. To address outlier sensitivity, FWSVD \cite{hsu2022languagemodelcompressionweighted} incorporates Fisher information, but this method requires gradients from the training process, leading to high computational costs. Zhang et al. \cite{zhang2024lossless} align the decomposition error with the negative gradient to reduce the impact of compression without retraining. Subsequently, SVD-LLM \cite{wang2024svd} integrates truncation-aware data whitening to relate singular values directly to compression loss. ASVD \cite{yuan2025asvdactivationawaresingularvalue} evaluates the sensitivity of weight matrices under the activation distribution, selects critical channels, and minimizes compression error on those channels. Despite these advances, most methods focus on compressing each layer's weight matrix independently, failing to fully exploit the structural redundancies that could be shared across layers.

%-------------------------------------------------------------------------
\subsection{Parameter Sharing}
Parameter sharing significantly reduces a model's parameter count by reusing weights across multiple layers. The Universal Transformer \cite{dehghani2018universal} proposes sharing weights completely across encoder and decoder layers, while Subformer \cite{reid2021subformer} partitions parameters into attention and feed-forward groups and shares weights within each group. Most existing methods adopt a group-wise sharing strategy, dividing layers into several groups that use identical weights. Dynamic Tying \cite{hay2024dynamic} schemes attempt to discover sharing structures during training using reinforcement learning, but their computational cost is prohibitive for large-scale models.

Training-free strategies like FiPS \cite{2025learningparametersharingtensor} compress ViTs and LLMs by minimizing block-level reconstruction error, but sharing is still confined to adjacent blocks. Basis Sharing \cite{wang2024basis} extends this idea by representing adjacent layers with a shared set of basis vectors and coefficient vectors. Although these methods avoid retraining, their sharing strategies are limited to adjacent layers or within the same module group. They do not systematically explore sharing structures that span multiple layers, leaving the selection of sharing configurations dependent on heuristics choices.

\section{Approach}
% This chapter formalizes parameter sharing through a colored bipartite graph $\Omega$ and uses its automorphism group $\mathrm{Aut}(\Omega)$ to characterize the permutation invariances induced by sharing. We begin from single-layer sharing and generalize it to cross-layer sharing.

\subsection{Parameter Sharing as Graph Coloring}
\textbf{Single-layer sharing.} We begin by formalizing parameter sharing within a single layer using edge coloring on bipartite graphs. Consider a neural layer that maps input vector $x \in \mathbb{R}^N$ to output vector $y \in \mathbb{R}^M$. We represent the layer's connectivity structure as a bipartite graph 
\begin{equation}
\Omega = (\mathcal{N}, \mathcal{M}, E),
\end{equation}
where $\mathcal{N} = \{1, \ldots, N\}$ indexes the input neurons. $\mathcal{M} = \{1, \ldots, M\}$ indexes the output neurons. $E \subseteq \mathcal{N} \times \mathcal{M}$ specifies which input-output pairs are connected.

Parameter sharing is encoded via a \emph{color function} $\alpha_{\text{edge}}: E \to C$ that assigns each edge $(n,m) \in E$ to a color class $c \in C$. The key insight is that edges sharing the same color use the same learnable parameter $\theta_c \in \mathbb{R}$. This yields the layer computation:
\begin{small}
\begin{equation}
\phi_\Omega(x)_m = \sigma \Big(\sum_{(n,m) \in E} \theta_{\alpha_{\text{edge}}(n,m)} \cdot x_n\Big), \quad m \in \mathcal{M},
\end{equation}
\end{small}
where $\sigma$ is the activation function, and $\theta_{\alpha_{\text{edge}}(n,m)}$ denotes the shared parameter assigned to edge $(n,m)$.

This framework captures various sharing patterns: no sharing (distinct colors), full sharing (uniform color), or structured repetition (e.g., convolution with local color patterns). Different colorings $\alpha_{\text{edge}}$ induce different sharing structures while preserving computational topology.

\textbf{Cross-layer sharing.} In large-scale models, similar transformation patterns often recur across layers, motivating a layer-wise form of parameter sharing.
Instead of coloring individual edges within a layer, we now color entire layers, encoding which layers share structural components.

To generalize this idea to cross-layer sharing, we color not “individual edges” but the “weight structures of entire layers.” We abstract the network’s layer structure as a higher-level bipartite graph whose nodes no longer correspond to input/output neurons, but to:
\begin{itemize}
    \item The set of layer indices — $\mathcal{L} = \{1,2,\dots,L\}$;
     \item A set of shared bases — $\mathcal{B} = \{B_1,B_2,\dots,B_K\}$. Each edge $(\ell,b)$ indicates that “layer $\ell$ uses shared basis $\mathcal{B}_b$.”
     \end{itemize}
     
To generalize the graph coloring idea to cross-layer sharing, we cannot simply make the weights of different layers exactly equal (i.e., $W_i = W_j$), as this is too restrictive and would harm model performance.
A more flexible solution is to share the structural properties of the weight matrix rather than all its parameters. We employ decomposition techniques. A shared basis $B_b$ is defined as a pair of factor matrices $B_b = (U_b, V_b)$, where $U_b \in \mathbb{R}^{M \times r}$ and $V_b \in \mathbb{R}^{r \times N}$ are the shared left and right factors, $r$ is the rank of the basis, trading off compression and expressivity; once layer $\ell$ selects basis $B_b$, it constructs its weight via a layer-specific coefficient $S_{\ell,b} \in \mathbb{R}^{r \times r}$ as $W_\ell \approx U_b S_{\ell,b} V_b^T.$

From this cross-layer perspective, it is more convenient to express 'which layer uses which basis' as a function:
\begin{small}
\begin{equation}
\begin{aligned}
\alpha_{\text{layer}}:\mathcal{L}\rightarrow \mathcal{B}, \quad \alpha_{\text{layer}}(\ell) = B_b
\end{aligned}\label{eq3}
\end{equation}
\end{small}
This function represents "the shared basis $B_b$ selected by layer $\ell$". Equivalently, this choice can also be represented using an indicator function,
\begin{small}
\begin{equation}
\begin{aligned}
A_{\ell,b} = \mathbb{I}[\alpha_{\text{layer}}(\ell) = B_b],
\end{aligned}\label{eq4}
\end{equation}
\end{small}
Thus forming a bipartite graph
$\Omega = (\mathcal{L}, \mathcal{B}, E_{\text{layer}}),$ where
\begin{small}
\begin{equation}
\begin{aligned}
E_{\text{layer}} = \{(\ell, \alpha_{\text{layer}}(\ell)) \mid \ell \in \mathcal{L}\}.
\end{aligned}\label{eq5}
\end{equation}
\end{small}

\textbf{Structural invariance and automorphism group.} When the shared structure is fixed by the mapping $\alpha_{\text{layer}}$, any layer permutation that keeps this mapping unchanged (i.e., after the permutation each layer still corresponds to the same shared basis) will not alter the parameterization form or the overall output of the network \cite{ravanbakhsh2017equivariance}. Intuitively, such permutations correspond to swapping layers that use the same basis. Formalizing this semantic: the automorphism group is defined as
\begin{small}
\begin{equation}
\begin{aligned}
\mathrm{Aut}(\Omega_{\text{layer}}) = \{\pi \in S_{\mathcal{L}} \mid \alpha_{\text{layer}}(\pi(\ell)) = \alpha_{\text{layer}}(\ell), \ \forall \ell \in \mathcal{L}\}
\end{aligned}\label{eq6}
\end{equation}
\end{small}

In other words, $\mathrm{Aut}(\Omega_{\text{layer}})$ consists of all layer index permutations that are interchangeable under the sharing structure. The shared bases induce a partition of the layer set $\mathcal{L}$ into color classes, each containing the layers sharing the same basis. 
If layers are divided by the choice of basis into color categories: 
\begin{small}
\begin{equation}
\begin{aligned}
\mathcal{L}_b = \{\ell \in \mathcal{L} \mid \alpha_{\text{layer}}(\ell) = B_b\}
\end{aligned}\label{eq7}
\end{equation}
\end{small}
Each layer within the same color class can be permuted arbitrarily, and the group structure decomposes into the direct product of the permutation groups of each color class. This implies that, under the cross-layer sharing structure, the model is invariant to any reordering of layers within the same color class, while swapping layers across different color classes would break this structural consistency.
For instance, if $L = 5$ and the mapping is
$\alpha_{\text{layer}} = (B_a, B_b, B_a, B_c, B_a)$,
then the color classes are
$\mathcal{L}_a = {1,3,5}$, $\mathcal{L}_b = {2}$, $\mathcal{L}_c = {4}$,
and the resulting automorphism group is
$\mathrm{Aut}(\Omega_{\text{layer}}) = S_3 \times S_1 \times S_1,$
which means that layers $1$, $3$, and $5$ can be freely permuted without changing the model output, while layers $2$ and $4$ are fixed.

The color function $\alpha_{\text{layer}}$ specifies which layers share a common basis, but it is not predetermined—it represents the core design variable of the sharing structure.
Finding the optimal coloring that balances compression and performance is a combinatorial problem and thus computationally intractable.
To address this, we introduce a surrogate objective with an efficient optimization algorithm to discover high-quality sharing configurations.

\subsection{Geometric Hessian-based Layer Coloring}
\textbf{Geometric Principle.} We now need to determine $\alpha_{\text{layer}}$, i.e., to choose a sharing scheme. To evaluate candidate colorings, we express the loss increase due to sharing by a local quadratic approximation and use it as the objective.

Let $W$ denote the original (non-shared) parameters, written per layer as $W = \{W_\ell\}_{\ell \in \mathcal{L}}$. For a candidate coloring $\alpha_{\text{layer}}$, denote the shared approximation by $\widehat{W}(\alpha_{\text{layer}})$. Define the parameter perturbation introduced by sharing as
$\delta := \widehat{W}(\alpha_{\text{layer}}) - W$.
Let $\mathcal{J}(\cdot)$ denote the model loss. The change in objective due to sharing is
\begin{small}
\begin{equation}
\Delta \mathcal{J} := \mathcal{J}\big(\widehat{W}(\alpha_{\text{layer}})\big) - \mathcal{J}(W).
\end{equation}
\end{small}
Expanding $\mathcal{J}$ around $W$ to second order and denoting $H = \nabla^2 \mathcal{J}(W)$ the Hessian at $W$, we obtain
\begin{small}
\begin{equation}
\Delta \mathcal{J} \approx \nabla \mathcal{J}(W)^\top \delta + \tfrac{1}{2}\delta^\top H \delta \approx \tfrac{1}{2}\delta^\top H \delta,\label{eq9}
\end{equation}
\end{small}
i.e. the linear term is negligible and the quadratic term dominates. Hence, for a given coloring $\alpha_{\text{layer}}$, the induced loss can be approximated by the quadratic cost $\tfrac{1}{2}\delta^\top H \delta$.

The quadratic cost $\tfrac{1}{2}\delta^\top H \delta$ describes a local "energy" in parameter space. Since $H$ near a local minimum is symmetric positive definite, we take its spectral decomposition
\begin{small}
\begin{equation}
H = Q \Lambda Q^\top,
\end{equation}
\end{small}
where $Q = [q_1, \dots, q_n]$ is orthogonal and $\Lambda = \text{diag}(\lambda_1, \dots, \lambda_n)$ with $\lambda_1 \le \lambda_2 \le \dots \le \lambda_n$. With the change of coordinates $z = Q^\top \delta$, the quadratic becomes
\begin{small}
\begin{equation}
\tfrac{1}{2}\delta^\top H \delta = \tfrac{1}{2} z^\top \Lambda z = \tfrac{1}{2}\sum_{i=1}^n \lambda_i z_i^2.
\end{equation}
\end{small}
Geometrically, the level set $\{\delta : \tfrac{1}{2}\delta^\top H\delta = c\}$ is an ellipsoid in $z$-space:
\begin{small}
\begin{equation}
\sum_{i=1}^n \lambda_i z_i^2 = 2c.
\end{equation}
\end{small}
The ellipsoid's major axes correspond to small eigenvalues: perturbations along these low-curvature directions increase the loss least per unit norm. Therefore, for a given perturbation magnitude, allocating perturbation energy to these directions is advantageous.
% \begin{algorithm}[H]
% \caption{Geometric Major-Axis Alignment Algorithm}
% \begin{algorithmic}[1]
% \REQUIRE Weights $\{W_\ell\}_{\ell=1}^L$, candidate bases $\{B_b=(U_b,\cdot,V_b)\}_{b=1}^K$, minor-axis count $t$, amplitude factor $\beta$
% \ENSURE Assignment $\alpha_{\rm layer}$
% \STATE Precompute for each layer $\ell$ orthonormal minor-axis vectors $\{p_j^{(\ell)}\}_{j=1}^t$
% \FOR{each layer $\ell$ and basis $B_b$}
%   \STATE $\delta \leftarrow U_b S_\ell V_b^\top - W_\ell$
%   \STATE $\delta^\parallel \leftarrow \delta - \sum_{j=1}^t \langle p_j^{(\ell)},\delta\rangle p_j^{(\ell)}$
%   \STATE $\tilde\delta \leftarrow$ clip$_{\tau_\ell}(\delta^\parallel)$ with $\tau_\ell=\beta\|W_\ell\|_F$
%   \STATE $s_\ell(B_b)\leftarrow \|\tilde\delta\|_2^2$
% \ENDFOR
% \STATE $\alpha_{\rm layer}(\ell)\leftarrow\arg\max_b s_\ell(B_b)\quad\forall\ell$
% \STATE $\widehat W_\ell \leftarrow U_{\alpha_{\rm layer}(\ell)} S_\ell V_{\alpha_{\rm layer}(\ell)}^\top$
% \RETURN $\alpha_{\rm layer},\{\widehat W_\ell\}$
% \end{algorithmic}
% \end{algorithm}

\textbf{Geometric Alignment Algorithm.} 
Intuitively, our goal is not to uniformly minimize the perturbation norm, but to encourage the sharing-induced difference $\delta$ to reside in the flattest region of the loss landscape---i.e., directions along which the objective is least sensitive.

To achieve this, we first focus on the high-curvature directions of the loss. Formally, let $\{p_j\}_{j=1}^{t}$ be an orthonormal basis of eigenvectors corresponding to the $t$ largest eigenvalues of the layer-wise Hessian. These vectors represent the \textbf{minor-axis directions}, where the loss is most sensitive. Any perturbation $\delta$ can be decomposed into its projection onto the high-curvature subspace  $\delta^\perp$, and its projection onto the low-curvature subspace (the major-axis directions)  $\delta^\parallel$:
\begin{equation} \label{eq:decomp_revised}
\delta = \delta^\parallel + \delta^\perp, \quad
\delta^\perp = \sum_{j=1}^{t}\langle p_j, \delta\rangle p_j,\quad \delta^\parallel = \delta - \delta^\perp
\end{equation}

Under a local quadratic approximation, the change in loss can be written as:
\begin{small}
\begin{equation} \label{eq:loss_approx_revised}
\delta^\mathsf{T}H\delta = \left( (\delta^\parallel)^\mathsf{T}H\delta^\parallel + (\delta^\perp)^\mathsf{T}H\delta^\perp \right).
\end{equation}
\end{small}
Since $\delta^\perp$ lies in the high-curvature subspace, its contribution to the loss is dominant. Therefore, \textbf{controlling the magnitude of the high-curvature component $\delta^\perp$ is key to maintaining model accuracy}.

Based on this, we propose the Geometric Alignment Principle: while allowing the low-curvature component $\delta^\parallel$ to vary within a defined region to preserve expressive power, we strictly minimize the energy of its high-curvature component $\delta^\perp$. This guides the total perturbation $\delta$ to align with the \textbf{major axes} of the Hessian ellipsoid. This principle is formulated as the following optimization problem:

\begin{equation} \label{eq:optim_goal_revised}
\min_{B_b} \|\delta^\perp(B_b)\|_2^2, \quad \text{s.t.} \quad \|\delta^\parallel(B_b)\|_2 \le \tau_\ell.
\end{equation}

Here, $B_b=(U_b, \cdot, V_b)$ is a candidate shared basis, and $\tau_\ell = \beta\|W_\ell\|_F$ is a trust-region radius, which bounds the magnitude of the low-curvature perturbation proportionally to the layer's weight norm. This formulation constrains the perturbation while naturally encouraging alignment with low-curvature directions. Moreover, since all layers sharing the same basis $B_b$ use identical transformations, this construction inherently ensures structural consistency without needing extra symmetry terms.

\begin{algorithm}[H]
\caption{Geometric Major-Axis Alignment}
\begin{algorithmic}[1]
\REQUIRE Weights $\{W_\ell\}_{\ell=1}^L$, candidate bases $\{B_b=(U_b,V_b)\}_{b=1}^K$, 
minor-axis count $t$, amplitude factor $\beta$.
\ENSURE Assignment $\alpha_{\rm layer}$, Aligned weights $\{\widehat W_\ell\}$.

\STATE Precompute for layer $\ell$ its orthonormal minor-axis vectors $\{p_j^{(\ell)}\}_{j=1}^t$.
\FOR{each layer $\ell$}
  \STATE // Select basis with minimal high-curvature energy
  \STATE $b^* \leftarrow \arg\min_b \left\| \sum_{j=1}^t \langle p_j^{(\ell)}, U_b S_{\ell,b} V_b^\top - W_\ell \rangle p_j^{(\ell)} \right\|_2^2$
  \STATE $\alpha_{\rm layer}(\ell) \leftarrow b^*$
  \STATE // Construct aligned weight using the selected basis
  \STATE $\delta^* \leftarrow U_{b^*} S_\ell V_{b^*}^\top - W_\ell$
  \STATE $\delta^{*\parallel} \leftarrow \delta^* - \sum_{j=1}^t \langle p_j^{(\ell)}, \delta^* \rangle p_j^{(\ell)}$
  \STATE $\tau_\ell \leftarrow \beta\|W_\ell\|_F$
  \STATE $\bar{\delta}^{*\parallel} \leftarrow \text{clip}_{\tau_\ell}(\delta^{*\parallel})$
  \STATE $\widehat W_\ell \leftarrow W_\ell + \bar{\delta}^{*\parallel}$
\ENDFOR
\RETURN $\alpha_{\rm layer}, \{\widehat W_\ell\}$
\end{algorithmic} \label{alg}
\end{algorithm}

The optimization is implemented via the following \textit{major-axis alignment procedure}. For each layer, we first estimate its local minor-axis basis $\{p_j\}$. Then, for each candidate basis $B_b$, we compute the induced perturbation $\delta$ and its high-curvature energy. The basis that minimizes this energy is selected:
\begin{equation} \label{eq:argmin_alpha_revised}
\alpha_{\text{layer}}(\ell) = \arg\min_b \|\delta^\perp(B_b)\|_2^2.
\end{equation}
After identifying the optimal basis, we construct the final aligned weights $\widehat{W}_\ell$ by discarding the high-curvature component and clipping the low-curvature component. Specifically, we compute the optimal perturbation $\delta^*$ and its low-curvature part $\delta^{*\parallel}$, then apply clipping to satisfy the constraint from Eq.~\eqref{eq:optim_goal_revised}. The final weight is updated as:
\begin{small}
\begin{equation} \label{eq:weight_update_revised}
\bar{\delta}^{*\parallel} = \text{clip}_{\tau_\ell}(\delta^{*\parallel}), \quad \text{and} \quad \widehat{W}_\ell = W_\ell + \bar{\delta}^{*\parallel}.
\end{equation}
\end{small}
This ensures the final weights are obtained via a perturbation that is strictly confined to the low-curvature subspace and has a controlled magnitude, thus maximizing performance preservation under weight sharing. The complete procedure is summarized in Algorithm~\ref{alg}. Specifically, each candidate basis is obtained from SVD-LLM, where every basis is inherently low-rank and thus directly controls the overall compression ratio.

\section{Experiments}
\subsection{Models and Datasets.}
We comprehensively evaluate our method across a diverse range of models. For visual Transformers, we evaluate the Swin-Transformer \cite{liu2021swin}  on ImageNet \cite{krizhevsky2012imagenet} and transfer it to downstream tasks such as CIFAR \cite{krizhevsky2009learning}. For large language models (LLMs), we conduct experiments on multiple architectures, including the LLaMA \cite{touvron2023llama} family (LLaMA-7B, LLaMA-13B, LLaMA-30B, LLaMA2-7B), OPT-6.7B \cite{zhang2022opt}, and Mistral-7B. Our evaluation encompasses 3 language modeling datasets: WikiText-2 \cite{merity2016pointer}, Penn Treebank (PTB) \cite{marcinkiewicz1994building}, and C4 \cite{raffel2020exploring}. Additionally, we assess performance on seven reasoning datasets: OpenbookQA \cite{banerjee2019careful}, WinoGrande \cite{sakaguchi2021winogrande}, HellaSwag \cite{zellers2019hellaswag}, PIQA \cite{bisk2020piqa}, MathQA \cite{amini2019mathqa}, ARC-easy, and ARC-challenge \cite{clark2018think}. All reasoning tasks are evaluated under zero-shot settings using the LM-Evaluation-Harness framework to ensure consistent and reproducible results.

\begin{table*}[ht]
\caption{Our method's PPL (↓) and zero-shot (↑) performance under LLaMA-7B, following an SVD-based evaluation scheme on 3 language modeling datasets and 7 common-sense reasoning datasets(\%). Ratio represents the compression rate.}
\renewcommand{\arraystretch}{1.2}
\setlength{\tabcolsep}{5.8pt}
\scalebox{0.83}{
\begin{tabular}{ccccc|cccccccc}
\hline
\textbf{Ratio} & \textbf{Method} & \textbf{PTB↓} & \textbf{C4↓} & \textbf{WikiText-2↓} & \textbf{Openb.} & \textbf{ARC\_e} & \textbf{WinoG.} & \textbf{HellaS.} & \textbf{ARC\_c} & \textbf{PIQA} & \textbf{MathQA} & \textbf{Average↑} \\ \hline
\multicolumn{1}{c|}{\color[HTML]{656565}0\%} & \color[HTML]{656565}Original & \color[HTML]{656565}8.35 & \color[HTML]{656565}7.34 & \color[HTML]{656565}5.68 & \color[HTML]{656565}28.0 & \color[HTML]{656565}67.0 & \color[HTML]{656565}67.0 &\color[HTML]{656565} 56.0 &\color[HTML]{656565} 38.0 &\color[HTML]{656565} 78.0 &\color[HTML]{656565} 27.0 &\color[HTML]{656565} 52.0 \\ \hline
\multicolumn{1}{c|}{} & SVD & 20306 & 18800 & 20061 & 14.0 & 27.0 & 51.0 & 26.0 & 21.0 & 53.0 & 21.0 & 31.0 \\
\multicolumn{1}{c|}{} & FWSVD & 2152 & 1511 & 1727 & 15.0 & 31.0 & 50.0 & 26.0 & 23.0 & 56.0 & 21.0 & 32.0 \\
\multicolumn{1}{c|}{} & ASVD & 16.55 & 15.93 & 11.14 & 25.0 & 53.0 & 64.0 & 41.0 & 27.0 & 68.0 & 24.0 & 43.0 \\
\multicolumn{1}{c|}{} & SVD-LLM & 18.05 & 15.93 & 7.94 & 22.0 & 58.0 & 63.0 & 43.0 & 29.0 & 69.0 & 24.0 & 44.0 \\
\multicolumn{1}{c|}{} & Basis Sharing & 17.35 & 15.03 & 7.74 & 28.0 & 66.0 & 66.0 & 46.0 & 36.0 & 71.0 & 25.0 & 48.0 \\
\multicolumn{1}{c|}{\multirow{-6}{*}{20\%}} & \cellcolor[HTML]{fde0fe}\textbf{Ours} & \cellcolor[HTML]{fde0fe}\textbf{16.54} & \cellcolor[HTML]{fde0fe}\textbf{13.88} & \cellcolor[HTML]{fde0fe}\textbf{7.07} & \cellcolor[HTML]{fde0fe}\textbf{29.0} & \cellcolor[HTML]{fde0fe}\textbf{66.1} & \cellcolor[HTML]{fde0fe}\textbf{68.5} & \cellcolor[HTML]{fde0fe}\textbf{46.4} & \cellcolor[HTML]{fde0fe}\textbf{37.4} & \cellcolor[HTML]{fde0fe}\textbf{71.1} & \cellcolor[HTML]{fde0fe}\textbf{25.1} & \cellcolor[HTML]{fde0fe}\textbf{49.1} \\ \hline
\multicolumn{1}{c|}{} & SVD & 17210 & 20871 & 13103 & 13.0 & 26.0 & 51.0 & 26.0 & 21.0 & 54.0 & 22.0 & 30.0 \\
\multicolumn{1}{c|}{} & FWSVD & 11058 & 7240 & 20127 & 17.0 & 26.0 & 49.0 & 26.0 & 22.0 & 51.0 & 19.0 & 30.0 \\
\multicolumn{1}{c|}{} & ASVD & 70 & 41 & 51 & 18.0 & 43.0 & 53.0 & 37.0 & 25.0 & 65.0 & 21.0 & 38.0 \\
\multicolumn{1}{c|}{} & SVD-LLM & 29.44 & 25.11 & 9.56 & 20.0 & 48.0 & 59.0 & 40.0 & 26.0 & 65.0 & 22.0 & 40.0 \\
\multicolumn{1}{c|}{} & Basis Sharing & 29.12 & 22.46 & 9.25 & 27.0 & 63.0 & 63.0 & 40.0 & 30.0 & 68.0 & 24.0 & 45.0 \\
\multicolumn{1}{c|}{\multirow{-6}{*}{30\%}} & \cellcolor[HTML]{fde0fe}\textbf{Ours} & \cellcolor[HTML]{fde0fe}\textbf{27.65} & \cellcolor[HTML]{fde0fe}\textbf{21.89} & \cellcolor[HTML]{fde0fe}\textbf{9.13} & \cellcolor[HTML]{fde0fe}\textbf{28.1} & \cellcolor[HTML]{fde0fe}\textbf{64.5} & \cellcolor[HTML]{fde0fe}\textbf{65.8} & \cellcolor[HTML]{fde0fe}\textbf{41.3} & \cellcolor[HTML]{fde0fe}\textbf{33.0} & \cellcolor[HTML]{fde0fe}\textbf{68.9} & \cellcolor[HTML]{fde0fe}\textbf{24.3} & \cellcolor[HTML]{fde0fe}\textbf{46.7} \\ \hline
\multicolumn{1}{c|}{} & SVD & 59977 & 47774 & 52489 & 15.0 & 26.0 & 52.0 & 26.0 & 22.0 & 53.0 & 20.0 & 30.0 \\
\multicolumn{1}{c|}{} & FWSVD & 20990 & 12847 & 18156 & 16.0 & 26.0 & 51.0 & 26.0 & 22.0 & 53.0 & 21.0 & 30.0 \\
\multicolumn{1}{c|}{} & ASVD & 3292 & 1109 & 1407 & 13.0 & 28.0 & 48.0 & 26.0 & 22.0 & 55.0 & 19.0 & 30.0 \\
\multicolumn{1}{c|}{} & SVD-LLM & 63.75 & 49.83 & 13.11 & 19.0 & 42.0 & 58.0 & 33.0 & 25.0 & 60.0 & 21.0 & 37.0 \\
\multicolumn{1}{c|}{} & Basis Sharing & 55.78 & 41.28 & 12.39 & 22.0 & 52.0 & 61.0 & 35.0 & 27.0 & 62.0 & \textbf{23.0} & 40.0 \\
\multicolumn{1}{c|}{\multirow{-6}{*}{40\%}} & \cellcolor[HTML]{fde0fe}\textbf{Ours} & \cellcolor[HTML]{fde0fe}\textbf{52.47} & \cellcolor[HTML]{fde0fe}\textbf{39.78} & \cellcolor[HTML]{fde0fe}\textbf{12.16} & \cellcolor[HTML]{fde0fe}\textbf{23.4} & \cellcolor[HTML]{fde0fe}\textbf{54.9} & \cellcolor[HTML]{fde0fe}\textbf{62.4} & \cellcolor[HTML]{fde0fe}\textbf{35.6} & \cellcolor[HTML]{fde0fe}\textbf{28.4} & \cellcolor[HTML]{fde0fe}\textbf{64.3} & \cellcolor[HTML]{fde0fe}\textbf{23.0} & \cellcolor[HTML]{fde0fe}\textbf{41.8} \\ \hline
\multicolumn{1}{c|}{} & SVD & 87227 & 79815 & 131715 & 16.0 & 26.0 & 50.0 & 26.0 & 23.0 & 52.0 & 19.0 & 30.0 \\
\multicolumn{1}{c|}{} & FWSVD & 28321 & 23104 & 24391 & 12.0 & 26.0 & 50.0 & 26.0 & 23.0 & 53.0 & 20.0 & 30.0 \\
\multicolumn{1}{c|}{} & ASVD & 47690 & 27925 & 15358 & 12.0 & 26.0 & 51.0 & 26.0 & 22.0 & 52.0 & 19.0 & 30.0 \\
\multicolumn{1}{c|}{} & SVD-LLM & 150.58 & 118.57 & 23.97 & 16.0 & 33.0 & 54.0 & 29.0 & 23.0 & 56.0 & 21.0 & 33.0 \\
\multicolumn{1}{c|}{} & Basis Sharing & 126.35 & 88.44 & 19.99 & 18.0 & 42.0 & 57.0 & \textbf{31.0} & 23.0 & 58.0 & 22.0 & 36.0 \\
\multicolumn{1}{c|}{\multirow{-6}{*}{50\%}} & \cellcolor[HTML]{fde0fe}\textbf{Ours} & \cellcolor[HTML]{fde0fe}\textbf{117.23} & \cellcolor[HTML]{fde0fe}\textbf{79.01} & \cellcolor[HTML]{fde0fe}\textbf{18.95} & \cellcolor[HTML]{fde0fe}\textbf{19.6} & \cellcolor[HTML]{fde0fe}\textbf{44.7} & \cellcolor[HTML]{fde0fe}\textbf{59.6} & \cellcolor[HTML]{fde0fe}\textbf{31.0} & \cellcolor[HTML]{fde0fe}\textbf{24.0} & \cellcolor[HTML]{fde0fe}\textbf{60.0} & \cellcolor[HTML]{fde0fe}\textbf{22.1} & \cellcolor[HTML]{fde0fe}\textbf{37.3} \\ \hline
\end{tabular}}
\end{table*}

\subsection{Implementation Details}
All models are implemented using Hugging Face transformers. LLaMA-30B is implemented in FP16 precision, while all other models use FP32. For sharing, we follow the Basis Sharing. All experiments are conducted on two NVIDIA A800 80GB GPUs. Second-order terms and eigenvectors are approximated using Hessian-vector products (HVP) combined with the Lanczos algorithm. The number of short-axis eigenvalues $t$ is set to 550, and the perturbation amplitude $\beta$ is set to 5e-2; these hyperparameter choices will be justified in the ablation studies. The clipping operation $\text{clip}_{\tau_\ell}$ refers to L2-norm clipping. All experimental code is implemented in PyTorch.

\subsection{Ablations}
\begin{figure}[ht]
\centering
\includegraphics[width=8.3cm,height=3cm]{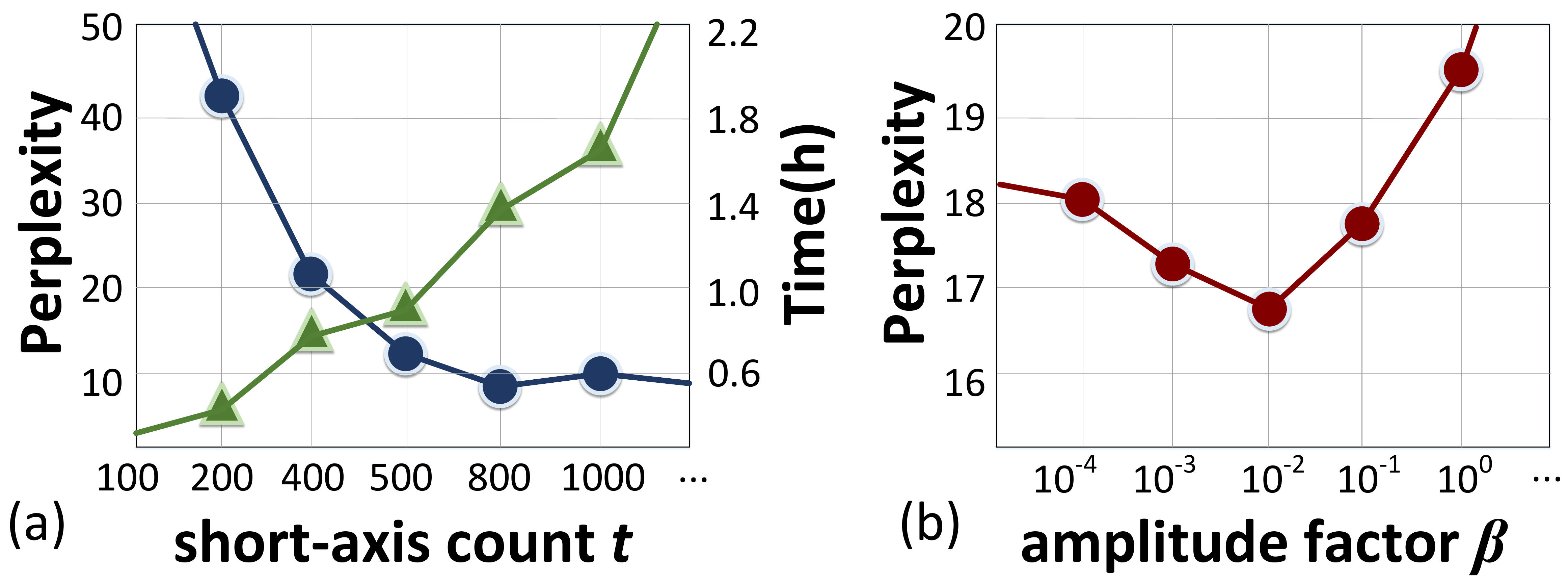}
\caption{Ablations. (a) As the number of minor axes increases, perplexity consistently decreases, though computational burden increases.
(b) When the amplitude factor increases, excessive perturbation leads to a sharp surge in perplexity.}
\vspace{-0.3cm}
\label{ablation}
\end{figure}

\textbf{The Number of Short-Axis.} Figure \ref{ablation}(a) examines the effect of the short-axis count $t$ in Algorithm. Increasing $t$ refines the estimation of high-curvature directions, allowing more accurate projection of perturbations onto the flat subspace. Consequently, perplexity decreases steadily, but computation time grows nearly linearly due to higher projection cost. This confirms that larger $t$ improves the curvature fidelity of the alignment but at the expense of efficiency.

\textbf{The Amplitude Factor.}
Figure \ref{ablation}(b) investigates the amplitude factor $\beta$, which controls the trust-region radius in alignment. Small $\beta$ overly restrict perturbations, preventing sufficient movement along flat directions and leading to underfitting. As $\beta$ increases to around $10^{-2}- 10^{-1}$, moderate perturbation energy improves alignment and yields the lowest perplexity. Beyond this range, excessive amplitude breaks the local quadratic assumption and injects noise into sensitive directions, sharply degrading performance.

\textbf{The First-Order Term.}
To verify the validity of the second-order approximation in Eq.\ref{eq9}, we evaluate the first-order contribution in the Taylor expansion on ViT.
We compute the ratio
$c = 2{\big|\nabla_W \mathcal{J}(W)^\top \delta\big|}/
{\big|\delta^\top H \delta\big|}$,
for each layer and each sampled perturbation \(\delta\) obtained from the sharing process.
Empirically, \(c<0.3\) for 90\% of layers, indicating that the first-order term is negligible compared with the second-order curvature term. This phenomenon arises because the model is already well optimized—the gradient norm \(\|\nabla_W \mathcal{J}\|\) is close to zero—making the first-order term vanish at convergence. Consequently, the loss change is dominated by the second-order, which validates the assumption used in Eq.\ref{eq9} and supports our sharing strategy.

\subsection{Comparison}
\textbf{Comparison on LLMs.} Across compression ratios, our method consistently outperforms all SVD-based baselines on both language modeling and zero-shot reasoning. Even under high compression, the model maintains stable perplexity and accuracy, indicating that the proposed major-axis alignment and perturbation control effectively preserve key representational structures. On LLaMA-7B, our approach achieves lower perplexity and higher average zero-shot scores across all datasets, showing clear advantages in both linguistic coherence and reasoning generalization. The gap further widens at higher compression, confirming that aligning shared subspaces with low-curvature Hessian directions improves robustness.

Similar trends appear on LLaMA2-7B in Figure \ref{llama2}, where our method achieves lower PPL than Basis-Sharing across all tested ratios. This suggests that the proposed geometric basis selection and adaptive amplitude modulation not only minimize distortion from rank reduction but also preserve global semantic consistency. These results demonstrate that Geo-sharing scales effectively to large models, preserving fluency and reasoning ability under strong compression.

\begin{table}[]
\caption{Comparison of our method's PPL (↓) performance on LLaMA2-7B with the baseline under different compression ratios.}
\renewcommand{\arraystretch}{1.2}
\setlength{\tabcolsep}{6.9pt}
\scalebox{0.93}{
\begin{tabular}{ccccc}
\hline
\textbf{Ratio}             & \textbf{Method}                       & \textbf{PTB↓}                           & \textbf{C4↓}                           & \textbf{WikiText-2↓}                   \\ \hline
{\color[HTML]{656565} 0\%} & {\color[HTML]{656565} Original}       & {\color[HTML]{656565} 7.29}             & {\color[HTML]{656565} 7.29}            & {\color[HTML]{656565} 5.47}            \\ \hline
                           & Basis Sharing                         & 60                                      & 15.3                                   & 7.77                                   \\
\multirow{-2}{*}{20\%}     & \cellcolor[HTML]{fde0fe}\textbf{Ours} & \cellcolor[HTML]{fde0fe}\textbf{54.53}  & \cellcolor[HTML]{fde0fe}\textbf{14.9}  & \cellcolor[HTML]{fde0fe}\textbf{7.57}  \\
                           & Basis Sharing                         & 97.4                                    & 23.86                                  & 9.69                                   \\
\multirow{-2}{*}{30\%}     & \cellcolor[HTML]{fde0fe}\textbf{Ours} & \cellcolor[HTML]{fde0fe}\textbf{88.33}  & \cellcolor[HTML]{fde0fe}\textbf{23.17} & \cellcolor[HTML]{fde0fe}\textbf{9.52}  \\
                           & Basis Sharing                         & 195.95                                  & 43.89                                  & 13.62                                  \\
\multirow{-2}{*}{40\%}     & \cellcolor[HTML]{fde0fe}\textbf{Ours} & \cellcolor[HTML]{fde0fe}\textbf{175.55} & \cellcolor[HTML]{fde0fe}\textbf{41.49} & \cellcolor[HTML]{fde0fe}\textbf{13.48} \\
                           & Basis Sharing                         & 509.3                                   & 98.92                                  & 21.3                                   \\
\multirow{-2}{*}{50\%}     & \cellcolor[HTML]{fde0fe}\textbf{Ours} & \cellcolor[HTML]{fde0fe}\textbf{371.75} & \cellcolor[HTML]{fde0fe}\textbf{88.27} & \cellcolor[HTML]{fde0fe}\textbf{20.16} \\ \hline
\end{tabular}} \label{llama2}
\end{table}

\begin{figure}[!ht]
\centering
\includegraphics[width=8.3cm,height=3.2cm]{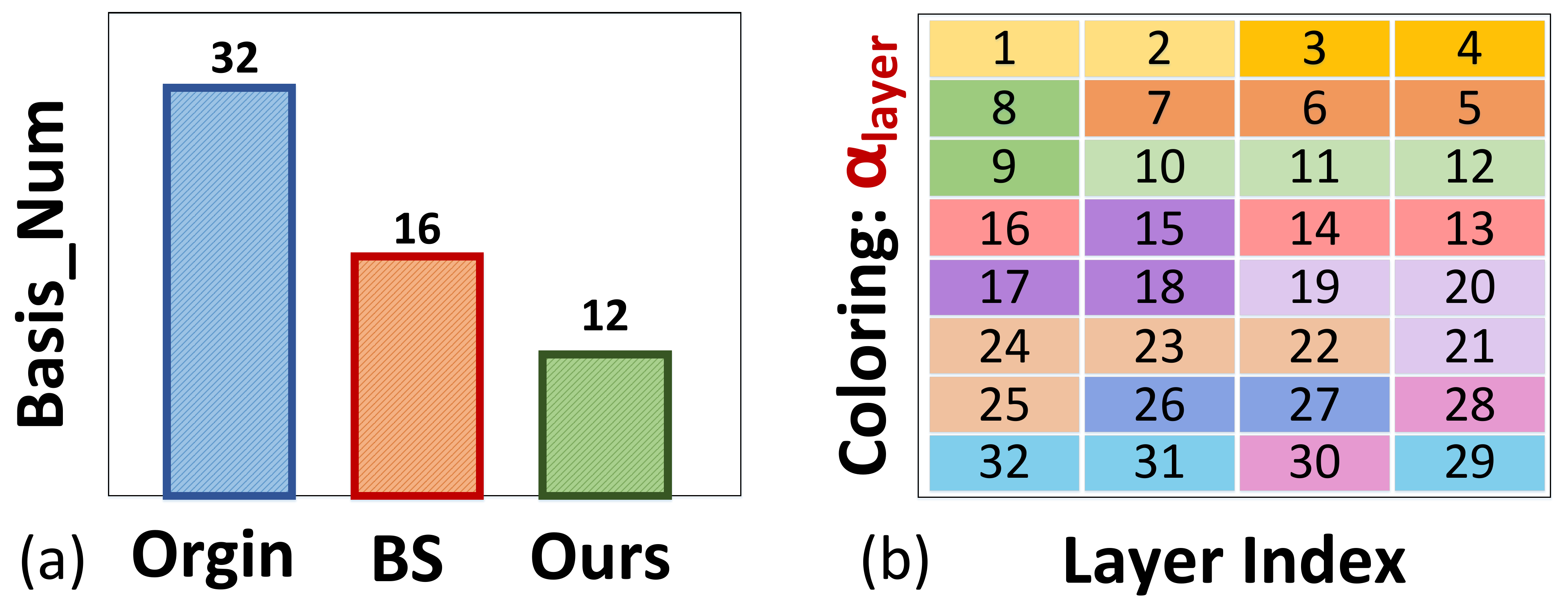}
\caption{(a) Comparison of the number of basis in our method (32 layers represented by only 12 basis) with existing methods \cite{wang2024basis}. (b) Specific coloring scheme $\alpha_{\text{layer}}$ of our method when compressing LLaMA 7B by 50\% (same color indicates shared basis).}
% \vspace{-0.3cm}
\label{group}
\end{figure}

Figure \ref{group} visualizes the layer-to-basis allocation under our group-constrained sharing framework. In Figure \ref{group}(a), our method represents a 32-layer network using only 12 bases, compared to 16 in Basis Sharing(BS) and 32 in the non-sharing case. This compact representation arises from enforcing layer permutations within the same equivalence class—an implicit automorphism group that governs basis sharing across layers. Figure \ref{group}(b) further shows the layer–basis mapping at a 50\% compression ratio on LLaMA-7B. Layers marked with the same color share 1 basis, forming groups that correspond to the invariant subsets under the automorphism action. This structured grouping confirms that our method effectively captures symmetry-induced redundancies, achieving high compression efficiency without breaking the model’s functional consistency.

\begin{table}[ht]
\caption{Performance comparison of our method and existing SVD-based approaches for Vision-Transformer on ImageNet.}
\renewcommand{\arraystretch}{1.2}
\setlength{\tabcolsep}{6.5pt}
\scalebox{0.82}{
\begin{tabular}{cccccc}
\hline
\textbf{Model} &  \textbf{Method} & \textbf{Top1}  & \textbf{Top1-Share} & \textbf{Top1-Drop} & \textbf{Ratio} \\ \hline
\multirow{5}{*}{Swin-L} &AAFM            & 86.25          & 85.73               & 0.52               & 25\%           \\
                        &GFM             & 86.25          & 85.83               & 0.42               & 25\%           \\
                        &FiPS            & 86.24          & 86.21               & 0.03               & 25\%           \\
                            &LossFac         & 86.23          & 86.19              & 0.04              & 27\%           \\
 &\cellcolor[HTML]{fde0fe} \textbf{Ours}   & \cellcolor[HTML]{fde0fe}\textbf{86.24} &\cellcolor[HTML]{fde0fe} \textbf{86.23}      &\cellcolor[HTML]{fde0fe} \textbf{0.01}      &\cellcolor[HTML]{fde0fe} \textbf{28\%}  \\ \hline
\multirow{2}{*}{DeiT-B} & FiPS            & 81.85            & 81.82               & \textbf{0.03}    & 25\%   \\
                        & \cellcolor[HTML]{fde0fe}\textbf{Ours}   & \cellcolor[HTML]{fde0fe}81.87            & \cellcolor[HTML]{fde0fe}81.84               &\cellcolor[HTML]{fde0fe} \textbf{0.03}    &\cellcolor[HTML]{fde0fe} 25\% \\  \hline  
\end{tabular}} \label{imagenet}
\end{table}

\textbf{Comparison on Vision transformer.}
Table \ref{imagenet} reports results on ImageNet using Swin-Transformer and DeiT backbones.
Compared with SVD-based baselines such as GFM \cite{yu2023compressing}, FiPS, and LossFac \cite{zhang2024lossless}, our method attains the smallest accuracy drop at a comparable compression ratio (28\%).
The post-sharing Top-1 accuracy remains nearly identical to the original model, showing that the proposed alignment mechanism effectively preserves representational capacity under strong compression.
These results confirm that Geo-Sharing generalizes well beyond language models, adapting effectively to vision backbones with hierarchical structures.

\begin{table}[ht]
\caption{Comparison of transfer learning results between multiple visual models and sharing model at different compression rates. Drop indicates the magnitude of performance drop.}
\renewcommand{\arraystretch}{1.2}
\setlength{\tabcolsep}{4.5pt}
\scalebox{0.82}{
\begin{tabular}{c|cc
>{\columncolor[HTML]{fde0fe}}c c
>{\columncolor[HTML]{fde0fe}}c c}
\hline
\textbf{}                    & \textbf{Model} & \textbf{Acc/Share(\%)} & \textbf{Drop} & \textbf{F1/Share(\%)} & \textbf{Drop} & \textbf{Ratio} \\ \hline
                                    & Swin-L         & 97.7/97.41         & -0.39           & 97.67/97.15      & -0.52          & 20\%             \\
                                    & Swin-B         & 90.81/91.00         & +0.20           & 90.77/91.44      & +0.67          & 20\%             \\
                                    & DeiT-B         & 92.90/91.90         & -1.00           & 92.87/91.90      & -0.97          & 30\%             \\
                                    % & Swin-L         & 97.7/97.33         & -0.27           & 97.67/97.28      & -0.39          & 30\%            \\
\multirow{-4}{*}{\rotatebox{90}{\textbf{CIFAR-10}}}  & Swin-B         & 90.81/91.36          & +0.45           & 90.77/91.35      & +0.58          & 30\%             \\ \hline
                                    & Swin-L         & 82.82/81.72        & -1.10           & 81.52/80.66      & -0.86          & 20\%             \\
                                    & Swin-B         & 67.39/68.05        & +0.66            & 65.35/66.06      & +0.71            & 20\%             \\
                                    % & Swin-L         & 82.82/82.47        & -0.35           & 81.52/81.29      & -0.23          & 30\%             \\
                                    & DeiT-B         & 72.80/70.30         & -2.50           & 71.33/69.56      & -1.78          & 30\%             \\
\multirow{-4}{*}{\rotatebox{90}{\textbf{CIFAR-100}}} & Swin-B         & 67.39/67.47       &   +0.08        & 65.35/65.14      & +0.21          & 30\%             \\ \hline
\end{tabular}} \label{transfer}
\end{table}

\textbf{Transferring Ability.}
In Table \ref{transfer}, we transfer the shared model to 2 downstream tasks, including CIFAR‑10/100. Consistent with the results on ImageNet, our method achieves accuracy on par with the original model \cite{dosovitskiy2020image, touvron2021trainingdataefficientimagetransformers} on these downstream tasks. This indicates that parameter sharing preserves the model’s generalization capability.

\begin{table}[h]
\centering
\caption{Inference efficiency of our method on real hardware.}
\renewcommand{\arraystretch}{1.2}
\setlength{\tabcolsep}{5.8pt}
\scalebox{0.74}{
\begin{tabular}{lcccc}
\hline
\textbf{Model} & \textbf{Params.(B)} & \textbf{MACs(B)} & \textbf{Latency(s)} & \textbf{Throughput(t/s)} \\ \hline
LLaMA2-7B & 6.74B & 6.61B & 13.21 & 1338.37 \\
\rowcolor[HTML]{fde0fe}
\textbf{Ours} & \textbf{3.50} \textcolor{teal}{\scriptsize{$\downarrow$48.1}\%} & \textbf{3.94} \textcolor{teal}{\scriptsize{$\downarrow$40.4}\%} & \textbf{7.06} \textcolor{teal}{\scriptsize{$\downarrow$46.6}\%} & \textbf{2152.92} \textcolor{magenta}{\scriptsize{$\uparrow$60.9}\%} \\ \hline
LLaMA-7B & 6.74B & 6.61B & 13.27 & 1331.88 \\
\rowcolor[HTML]{fde0fe}
\textbf{Ours} & \textbf{3.99}  \textcolor{teal}{\scriptsize{$\downarrow$40.8}\%} & \textbf{3.94} \textcolor{teal}{\scriptsize{$\downarrow$40.4}\%} & \textbf{7.38} \textcolor{teal}{\scriptsize{$\downarrow$44.4}\%} & \textbf{2084.30} \textcolor{magenta}{\scriptsize{$\uparrow$56.5}\%} \\ \hline
Mistral-7B & 7.24B & 7.11B & 14.61 & 1248.48 \\
\rowcolor[HTML]{fde0fe}
\textbf{Ours} & \textbf{3.75} \textcolor{teal}{\scriptsize{$\downarrow$48.2}\%} & \textbf{3.99} \textcolor{teal}{\scriptsize{$\downarrow$43.9}\%} & \textbf{7.93} \textcolor{teal}{\scriptsize{$\downarrow$45.7}\%} & \textbf{2135.37} \textcolor{magenta}{\scriptsize{$\uparrow$71.0}\%} \\ \hline
\end{tabular}}
\label{tab:latency}
\end{table}

\textbf{Inference Efficiency on Real Hardware.} Table \ref{tab:latency} summarizes the inference efficiency of our method across three representative  models, evaluated on a single NVIDIA A800 GPU with batch size 512 and sequence length 32.
Across all models, our approach consistently reduces both parameter count and MACs by around 40–50\%, leading to nearly 45\% lower latency and up to 70\% higher throughput compared with the original model.
These results indicate that our method not only maintains model accuracy but also brings substantial runtime benefits, showing strong generalization across different architectures and demonstrating its practicality for large-scale deployment.

\textbf{Algorithm Time.} During selection, for a batch size of 1, a context length of 512, and 20 iterations, computing the eigenvalues (top 550 eigenvalues) takes approximately 0.93 hours, while the remaining high-curvature energy minimization requires 0.4 hours. In comparison, Dynamic Tying \cite{hay2024dynamic} take around 13.8 hours. Geo-Sharing demonstrates significantly faster efficiency while achieving lower PPL. During deployment, our method maintains lower PPL while keeping inference time comparable to Basis Sharing.

\begin{figure}[!ht]
\centering
\includegraphics[width=7.2cm,height=4.2cm]{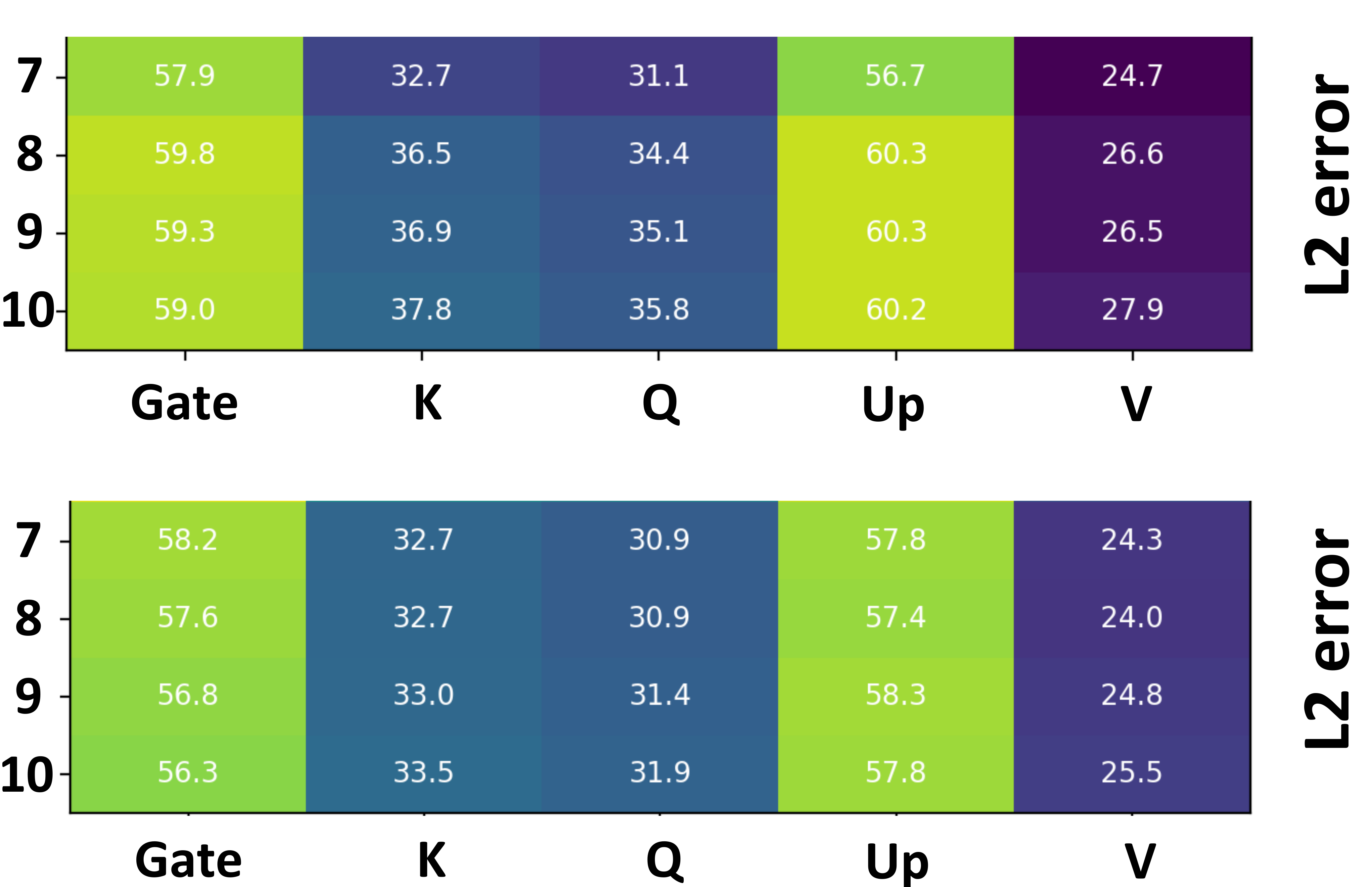}
\caption{Comparison of L2 error caused by different coloring functions (top: adjacent, bottom: Geo-sharing), including the difference between the weights after sharing in layers 7-10 and the weights of the original standard model.}
% \vspace{-0.3cm}
\label{proof}
\end{figure}

\textbf{Visualization of Axis-Aligned Validity.}
We further validate the effectiveness of the curvature-guided coloring function used in Geo-Sharing.
Figure \ref{proof} compares the L2 reconstruction error across layers 7–10 for different parameter groups (Gate, K, Q, Up, V) under two coloring strategies: the heuristic adjacent and our Geo-Sharing.
The Geo-Sharing coloring yields consistently lower errors across all modules, indicating that aligning sharing directions with the major (low-curvature) axes of the Hessian effectively reduces post-sharing distortion.
By grouping layers that occupy similar regions in the curvature space, the proposed coloring function also constructs transformation-consistent groups, thereby preserving the model’s layer-wise equivariance while improving compression fidelity.

\subsection{Scalability}

\begin{table}[]
\caption{Perplexity comparison between our method and Basis Sharing under extreme compression ratios on C4 and WikiText-2.}
\renewcommand{\arraystretch}{1.2}
\setlength{\tabcolsep}{13.8pt}
\scalebox{0.85}{
\begin{tabular}{cc
>{\columncolor[HTML]{fde0fe}}c c}
\hline
\textbf{Dataset} & \textbf{Basis Sharing} & \textbf{Ours}     & \textbf{Ratio}         \\ \hline
% C4               & 264.0663               & \textbf{228.7611} &                        \\
% Wikitext-2       & 46.8258                & \textbf{40.9742}  & \multirow{-2}{*}{60\%} \\ \hline
C4               & 651.8314               & \textbf{603.4069} &                        \\
WikiText-2       & 136.8194               & \textbf{125.0952} & \multirow{-2}{*}{70\%} \\ \hline
C4               & 2465.999               & \textbf{995.33}   &                        \\
WikiText-2       & 624.0834               & \textbf{424.8948} & \multirow{-2}{*}{80\%} \\ \hline
\end{tabular}}
\label{tabcompression}
\end{table}

\begin{table}[]
\caption{Scalability results for larger-scale LLMs on WikiText-2.}
\renewcommand{\arraystretch}{1.2}
\setlength{\tabcolsep}{8.0pt}
\scalebox{0.84}{
\begin{tabular}{cccc}
\hline
\textbf{Model} & \textbf{LLaMA-7B} & \textbf{LLaMA-13B} & \textbf{LLaMA-30B} \\ \hline
SVD            & 20061             & 946.31             & 54.11              \\
FWSVD          & 1630              & OOM                & OOM                \\
SVD-LLM        & 7.94              & 6.61               & 5.63               \\
Basis Sharing  & 7.74              & 6.51             & 5.47               \\
\rowcolor[HTML]{fde0fe} 
\textbf{Ours}  & \textbf{7.07}     & \textbf{6.21}      & \textbf{5.33}      \\ \hline
\end{tabular}}\label{huge}
\end{table}

\textbf{Larger-scale LLMs. }To verify adaptability on larger models, we extend experiments to LLaMA-7B, 13B, and 30B.
As shown in Table \ref{huge}, existing methods (FWSVD) fail to scale due to high memory cost(OOM), while ours achieves the lowest perplexity across all sizes.

\textbf{Extreme compression.} Under extreme compression ratios of 70–80\%, our method consistently outperforms Basis Sharing with lower perplexity on both C4 and WikiText-2 datasets (Table \ref{tabcompression}). Even at 80\% compression, it reduces perplexity by over 50\% on C4, showcasing the representational stability of the group-constrained basis formulation under severe axis reduction.

\begin{table}[]
\caption{Scalability results of LLMs with different structures.}
\renewcommand{\arraystretch}{1.2}
\setlength{\tabcolsep}{8.5pt}
\scalebox{0.86}{
\begin{tabular}{cccc}
\hline
\textbf{Method} & \textbf{LLaMA-7B} & \textbf{OPT-6.7B} & \textbf{Mistral-7B} \\ \hline
LLM-Pruner      & 19.09             & -                 & -                   \\
ASVD            & 11.14             & 82                & 10.21               \\
SVD-LLMv2       & 7.12              & 13.46             & -                   \\
Basis Sharing   & 7.74              & 11.79             & 7.57                \\
\rowcolor[HTML]{fde0fe} 
\textbf{Ours}   & \textbf{7.07}     & \textbf{11.68}    & \textbf{7.49}          \\ \hline
\end{tabular}} \label{otherllm}
\end{table}

\textbf{Diverse LLM Architectures.} As shown in Table \ref{otherllm}, our method achieves the lowest perplexity on WikiText-2 under a 20\% compression ratio across OPT-6.7B, LLaMA-7B, and Mistral-7B. This demonstrates strong scalability and generalization across LLMs without model-specific tuning.

For more related works, mathematical(Group Theory, Algebraic Geometry, Hessian) details and experiments on Geo-Sharing, please see the appendix.

\section{Conclusion}
This paper presents Geo-Sharing, a structured compression framework that reformulates multi-layer parameter sharing through geometric and symmetry principles.
By enforcing layer-wise symmetry via a coloring function and aligning shared subspaces with Hessian’s low-curvature directions, Geo-Sharing achieves efficient, theoretically grounded sharing configurations.
Experiments across vision and language models demonstrate superior compression–accuracy trade-offs over heuristic and SVD-based methods, providing a robust, interpretable, training-free solution for scalable model deployment.
{
    \small
    \bibliographystyle{ieeenat_fullname}
    \bibliography{main}
}


\end{document}